\documentclass[aps,physrev,reprint,groupedaddress]{revtex4-2}

\usepackage{amsthm,amsmath,amsfonts,amssymb}
\usepackage{hyperref}
\hypersetup{
    colorlinks=true,
    linkcolor=blue,
    filecolor=magenta,      
    urlcolor=cyan,
    citecolor=green,
    }
\usepackage{graphicx, wrapfig}
\usepackage{mathtools}
\usepackage{dsfont}
\usepackage{bm}
\usepackage{enumitem}
\usepackage{color,soul}
\usepackage{booktabs}
\usepackage{multirow}

\theoremstyle{plain}              
\newtheorem{theorem}{Theorem}
\newtheorem{corollary}[theorem]{Corollary}
\newtheorem{lemma}[theorem]{Lemma}

\newtheorem{remark}{Remark}
\newtheorem{definition}{Definition}
\newtheorem{example}{Example}
\def\lb{\label}

\newcommand{\defeq}{\vcentcolon=}

\DeclarePairedDelimiter{\abs}{\lvert}{\rvert}  

\DeclarePairedDelimiterX{\divx}[2]{(}{)}{%
  #1\;\delimsize\|\;#2%
}
\newcommand{\KL}{\mathrm{KL}\divx} 

\makeatletter
\newcommand{\@giventhatstar}[2]{\left(#1\;\middle|\;#2\right)}
\newcommand{\@giventhatnostar}[3][]{#1(#2\;#1|\;#3#1)}
\newcommand{\giventhat}{\@ifstar\@giventhatstar\@giventhatnostar}
\makeatother

\def\d{\mathrm{d}}

\def\ain{I^+}
\def\cain{I^\oplus}

\def\sfA{\mathsf{A}}

\def\sfT{\mathsf{T}}

\def\calF{\mathcal{F}}
\def\calG{\mathcal{G}}
\def\calH{\mathcal{H}}

\def\calU{\mathcal{U}}

\def\calX{\mathcal{X}}

\def\frakF{\mathfrak{F}}
\def\frakP{\mathfrak{P}}

\def\bbN{\mathbb{N}}
\def\bbR{\mathbb{R}}

\def\beq{\begin{equation}}
\def\eeq{\end{equation}}
\def\lb{\label}

\def\bfP{\mathbf P}
\def\bfQ{\mathbf Q}

\def\1{\mathds{1}} 

\def\bee{\begin{equation}}
\def\eeq{\end{equation}}
\def\lb{\label}

\begin{document}


\title{Conserved active information}


\author{Yanchen Chen}
\author{Daniel Andrés Díaz--Pachón}%
 \email{Contact author: DDiaz3@miami.edu}
\affiliation{%
 Division of Biostatistics, University of Miami, Miami, FL, USA.
}%


\date{\today}

\begin{abstract}
We introduce conserved active information, a symmetric extension of active information that quantifies net information gain or loss across an entire measurable space. Using Bernoulli examples, we show that conserved active information reveals regimes hidden by Kullback-Leibler divergence, such as the simultaneous reduction of local and global disorder. We formally prove these regimes in binary settings, distinguishing disorder-increasing reducible regimes from order-imposing emergent ones. We further illustrate these regimes with examples from Markov chains and cosmological fine-tuning. This resolves a longstanding question about active information, showing that, up to a threshold, positive active information (increased local order) can be produced by reshuffling the available information in the problem (decreased global order), without requiring exogenous information. However, beyond the threshold, information is indeed exogenous, forcing the system to be open.
\end{abstract}

\keywords{entropy, information, statistical mechanics}

\maketitle


\section{Introduction}

Information theory imposes and is subject to highly consequential limitations. Among the most striking is its self-imposed focus on averages (entropy, Kullback-Leibler divergence, mutual information, etc.). However, this approach is unsatisfactory when highly relevant and informative events are under consideration. For these events, informational averages underestimate the amount of information required to produce them. For instance, inspired by the small number of amino acid sequences that encode proteins, Nobel laureate Jack Szostak and collaborators defined functional configurations in a vast combinatorial space, as the sequences that achieve a minimal ``specified degree of function'' \cite{HazenEtAl2007, Szostak2003}. Other examples of relevant and informative events include, but are not limited to, the origin of the universe \cite{BarnesLewis2020, Luminet2024}, cosmological singularities \cite{HawkingPenrose1970, Penrose2004}, life-permitting intervals in cosmological fine-tuning \cite{Davies2008, LewisBarnes2016, Rees2000}, major evolutionary transitions \cite{Carter1983, Noble2017}, and targets in computational search problems \cite{MarksDembskiEwert2017}. Therefore, information theory needs to consider techniques based on local events and unaveraged measures that can inform scientific discovery.

In this paper, we introduce {\bf conserved active information} (CAI). CAI is based on {\bf active information} (AIN), which was proposed to quantify the information required to reach a target in a search problem \cite{DembskiMarks2009b}. The concept can be readily extended beyond search spaces to any probability space (Section II). We prove that CAI partitions systems with positive AIN into reducible and emergent regimes, where the latter renders it impossible to regard AIN as merely a redistribution of the system's available information, forcing us to see a system as open (Section III). Among others, this finding induces a very natural definition of {\it emergence} as a specification whose AIN is beyond the given threshold, thus removing the laxity with which the word is used in the literature \cite{Levin2025}.  

The paper is structured as follows: Section II defines active information and examines its key properties. Section III introduces CAI, presents results that demonstrate its properties, and illustrates its importance with examples. In Section IV, we conclude with a discussion of the relevance of CAI and some open problems.

\section{Active information}\lb{S:AIN}

Another way to see the limitations of averaging is through the highly consequential No Free Lunch theorems (NFLTs), which assert that, on average, no search outperforms a random one. The NFLTs imply that, ``if an algorithm performs better than random search on some class of problems, then it must perform {\it worse than random search} on the remaining problems'' \cite{WolpertMacReady1997}, underscoring ``the importance of incorporating problem-specific knowledge into the behavior of the algorithm'' \cite{WolpertMacReady1995, WolpertMacReady1997}. According to the NFLTs, it is precisely this problem-specific knowledge that enables biasing an algorithm to reach the target more effectively than a blind search (a fact that is particularly true in machine learning \cite{Montanez2017a, Wolpert2021}). Therefore, AIN $\ain$, was introduced to measure the amount of information $I_2$ infused by a programmer to reach a target $\sfT$ in a search space $\calX$, relative to the information $I_1$ provided by a baseline \cite{DembskiMarks2009a, DembskiMarks2009b, DembskiMarks2010, MarksDembskiEwert2017}: 
\begin{align}\lb{AIN}
\begin{aligned}
    \ain &= \ain_\sfT \divx{\bfP_1}{\bfP_2} = \ain\divx{\bfP_1}{\bfP_2}(\sfT) \\
    &\defeq I_1(\sfT) - I_2(\sfT) =  \log \frac{\bfP_2(\sfT)}{\bfP_1(\sfT)},
 \end{aligned}
\end{align}
where $I_i(\sfT) = -\log \bfP_i(\sfT)$ is the self-information of $\sfT$ under the law $\bfP_i$, for $i=1,2$, and we assume $\log(0/0)= 0$ by continuity. We allow $\ain$ to take values in the extended real line $\overline\bbR \defeq \bbR \cup \{\infty\} \cup \{-\infty\}$. As it is customary in search problems, we also assume $\abs{\sfT} \ll \abs{\calX} < \infty$. The probabilities $\bfP_1$ and $\bfP_2$ denote random search and search with additional knowledge, respectively. Thus, $I_1(\sfT)$ is the {\bf endogenous information}, measuring the baseline difficulty of finding $\sfT$, and $I_2(\sfT)$ is the {\bf exogenous information}, measuring the difficulty of the problem given the additional knowledge, making the AIN $\ain$ the problem-specific information incorporated into the search. 

Observe that in \eqref{AIN}, little more than $\sfT$ being measurable {\it by} the two underlying probability spaces is required. Formally, for two probability spaces $(\calX_1, \calF_1, \bfP_1)$ and $(\calX_2, \calF_2, \bfP_2)$, the only requirement is that $\sfT \in \calF_1 \cap \calF_2$. In applications, this distinction between the two spaces is key. For instance, in the Brillouin AIN, the programmer learns that the target $\sfT \subset \calX_2 \subset \calX_1$ lies within a subset of the original space \cite{Brillouin1956}. Then, assuming that $\bfP_1$ corresponds to a blind search, $\ain_\sfT = \log \bfP_2(\sfT) - \log \bfP_1(\sfT) = \log(\abs{\calX_1}/\abs{\calX_2})$. However, for theoretical purposes, it is better to work with a single measurable space $(\calX,\calF)$: \begin{lemma}\lb{UniSpace}
	Let $\calX = \calX_1 \cup \calX_2$, $\calF = \sigma(\calF_1, \calF_2)$, and extend the measures $\bfP_i$, for $i=1,2$, to
$$
	\bfP_i^*(\sfA) =
	\begin{cases}
		\bfP_i(\sfA) & \text{if } \sfA\in \calF_i,\\
		0 & \text{otherwise}.
	\end{cases}
$$
Then the measurable space $(\calX, \calF)$ and the probability measures $\bfP_1^*, \bfP_2^*$ associated with it exist and are well-defined.
\end{lemma}

\begin{proof}
    For $i=\{1,2\}$, we need to verify that 
    \begin{itemize}
        \item $\bfP_i^*(\calX) = 1$,
        \item $\bfP_i^*(\sfA) \ge 0$,
        \item Countable additivity: If $\{\sfA_j^*\}_{j\in \bbN} \subset \calF$ is a disjoint sequence of measurable sets, then $\bfP_i^*\left(\bigcup_j \sfA_j^*\right) = \sum_j \bfP_i^*(\sfA_j^*)$, for $i\in\{1,2\}$.
    \end{itemize}
    The first two properties are trivial. As for countable additivity, redefine $\sfA_j^*$ as
    \begin{align*}
        \sfA_j^* = 
        \begin{cases}
            \sfA_j & \mathrm{if\ } \sfA_j^* \in \calF_i \\
            \emptyset & \mathrm{ otherwise.}
        \end{cases}
    \end{align*}
    Then 
    \begin{align*}
    \bfP_i^*\left(\bigcup_j \sfA_j^*\right) = \bfP_i\left(\bigcup_j \sfA_j\right) = \sum_j \bfP_i(\sfA_j) = \sum_j \bfP_i^*(\sfA_j^*), 
    \end{align*}
    as required.
\end{proof}

\begin{remark}
    With a slight abuse of notation, we will still refer to $\bfP_1^*$ and $\bfP_2^*$ as $\bfP_1$ and $\bfP_2$, respectively.
\end{remark}

This unified space enables the generalization of active information to all probability spaces, not only the finite ones considered in search problems, while retaining the interpretation: $\ain$ is the excess of information required to produce $\sfT$ when $\bfP_2$ is used instead of some baseline $\bfP_1$. Indeed, the measure-theoretic formulation allows writing AIN as 
\begin{align}\lb{AIN2}
	\ain = \int_\sfT \log \frac{p_2(x)}{p_1(x)} \d \mu,
\end{align}
where $\mu$ is a measure on $(\calX, \calF)$, $\bfP_i$ is absolutely continuous w.r.t.~$\mu$ (noted $\bfP_i \ll \mu$), and $p_i = \d \bfP_i/\d \mu$ is its Radon--Nikodym derivative, for $i=1,2$ (such a measure $\mu$ always exists, for example, $\mu = (\bfP_1 + \bfP_2)/2$). Therefore, great generality is achieved, depending on the choice of $\calX$, $\calF$, $\bfP_2$, $\bfP_1$, $\sfT$, and $f$ (defined below):

\begin{definition}\lb{Spec}
	Let $f:\calX \to \bbR$ be a real-valued measurable function on $(\calX, \calF)$ and assume we are interested in events $\sfT$ where $f$ is large:
\begin{align}\label{T}
	\sfT \defeq \{x\in\calX; \, f(x)\ge f_0\} \in \calF
\end{align}
for some $f_0\in\bbR$. Then we say that $\sfT$ is a {\bf specification} and $f$ is a {\bf specification function}.
\end{definition}

We now define the {\bf Kullback-Leibler divergence} (KL):
\begin{align}\label{KL}
\begin{aligned}
	&\KL{\bfP_2}{\bfP_1} =\KL{X_2}{X_1} \\
    &\defeq 
	\begin{cases}
		\int p_2(x) \log\frac{p_2(x)}{p_1(x)} \mu(\d x) & \text{if } \bfP_1,\bfP_2 \ll \mu,\\
		\infty & \text{otherwise},
	\end{cases}
\end{aligned}
\end{align}
where $\mu$ is as in \eqref{AIN2}. Some have proposed KL as the most basic concept in information theory because it applies equally to discrete and continuous random variables, is always nonnegative and is uniquely defined, whereas entropy, when extended to continuous random variables, can be negative and is not invariant under transformation of the distribution's parameters \cite{Chodrow2019, Verdu2009}.

Returning to AIN in \eqref{AIN2}, it is a comparative measure that quantifies how much information is added or removed when viewing $\sfT$ through $\bfP_2$ rather than $\bfP_1$, even when the space is not discrete. To illustrate some properties of AIN, let $\frakP$ be the class of all probability measures on $(\calX,\calF)$. Then, for all $\bfP, \bfP^*, \bfP' \in \frakP$ and all $\sfA \in \calF$, AIN satisfies the following properties:
\begin{enumerate}[label=\textbf{P\arabic*}]
	\item\label{Prop1} Can be positive, negative, or zero:
    		\begin{itemize}
        			\item $\ain_\sfA\divx{\bfP}{\bfP^*} < 0$ iff $\bfP(\sfA) < \bfP^*(\sfA)$, 
        			\item $\ain_\sfA\divx{\bfP}{\bfP^*} = 0$ iff $\bfP(\sfA) = \bfP^*(\sfA)$, 
        			\item $\ain_\sfA\divx{\bfP}{\bfP^*} > 0$ iff $\bfP(\sfA) > \bfP^*(\sfA)$. 
    		\end{itemize}
	\item\label{Prop2} Induces a total ordering of $\frakP$, in the sense that $\bfP \ge \bfP^*$ iff $\ain_\sfA\divx{\bfP}{\bfP^*} \ge 0$. It satisfies		
		\begin{itemize}
			\item Reflexivity: $\ain_\sfA\divx{\bfP}{\bfP} \ge 0$ iff $\bfP(\sfA) \ge \bfP(\sfA)$.
			\item Antisymmetry: $\ain_\sfA\divx{\bfP}{\bfP'} \ge 0$ and $\ain_\sfA\divx{\bfP'}{\bfP} \ge 0$ iff $\bfP(\sfA) = \bfP'(\sfA)$.
			\item Transitivity: If $\ain_\sfA\divx{\bfP}{\bfP^*} \ge 0$ and $\ain_\sfA\divx{\bfP'}{\bfP} \ge 0$, then $\ain_\sfA						\divx{\bfP'}{\bfP^*} \ge 0$ and $\bfP'(\sfA) \ge \bfP^*(\sfA)$.
			\item Totality: $\ain_\sfA\divx{\bfP}{\bfP'} \ge 0$ or $\ain_\sfA\divx{\bfP'}{\bfP} \ge 0$.
		\end{itemize}
\end{enumerate}

Properties \ref{Prop1}--\ref{Prop2} establish other important differences between AIN and KL. First, KL cannot be negative; therefore, it is not a comparative measure and cannot quantify local information changes between $\bfP_2$ and $\bfP_1$. This is a key consideration in Section III. Second, KL does not induce any ordering whatsoever because, as a relation, it is neither antisymmetric nor transitive. The only comparison we can draw about $\bfP_2$ and $\bfP_1$ from $\KL{\bfP_2}{\bfP_1}$ is that $\KL{\bfP_2}{\bfP_1} = 0$ iff $\bfP_2 = \bfP_1$ a.s. Third, KL can be used to obtain AIN, but AIN cannot generally be obtained from KL. A third important property of AIN, shared by the KL divergence, is:
\begin{enumerate}[label=\textbf{P3}]
	\item\label{Prop3} AIN tensorizes. That is, for an $n$-dimensional Cartesian product $\sfA= \sfA_1 \times \cdots \times\sfA_n$, 		and product measures $\bfP_2 \defeq P_{21} \otimes \cdots \otimes P_{2n}$, $\bfP_1 \defeq P_{11} \otimes \cdots \otimes 		P_{1n}$, where $P_{21}, \ldots, P_{2n}$ and $P_{11}, \ldots, P_{1n}$ are independent, and $P_{11}(\sfA_1) \cdots 			P_{1m}(\sfA_n) \neq 0$, we have that $\log [\bfP_2(\sfA)/ \bfP_1(\sfA)] = \sum_{i=1}^n \log [P_{2i}(\sfA_i)/P_{1i}(\sfA_i)]$. 
\end{enumerate}

In high-dimensional scenarios, property \ref{Prop3} enables decomposition in product spaces. Moreover, \ref{Prop3} makes AIN preferable to 
\beq\lb{DiffP}
	\bfP_2(\sfA) - \bfP_1(\sfA)
\eeq 
because \eqref{DiffP} does not tensorize. Among other beneftis, this allows a more practical redefinition of prevalence bias as follows: If $\bfP_1(\sfA)$ is the true prevalence of $\sfA$ and $\bfP_2(\sfA)$ is the expected value of its estimator, $\ain$ is preferable to the usual definition of bias as \eqref{DiffP}; see, for instance, \cite{DiazEtAl2025, HossjerEtAl2024, MontanezEtAl2021, MontanezEtAl2019, ZhouEtAl2023}. Also, because of \ref{Prop3}, KL is usually preferable to the total variation distance $\mathrm{TV} = \mathrm{TV}(\bfP_2,\bfP_1) \defeq \sup_{\sfA \in \calF}|\bfP_2(\sfA) - \bfP_1(\sfA)|$ because, by Pinsker's inequality, TV can be approximated by KL:
\begin{align}\label{TV}
\mathrm{TV}(\bfP_2,\bfP_1) \le \sqrt{\KL{\bfP_2}{\bfP_1}/2}.
\end{align}

For analytical developments and diverse applications of specifications and AIN, see \cite{DiazEtAl2025, DiazHossjer2022, DiazHossjerMarks2021, DiazHossjerMarks2023, DiazHossjerMathew2024, LiuDiazRao2026, DiazMarks2020b, DiazMarks2020a, DiazRao2021, DiazSaenzRao2020, DiazEtAl2019, HossjerEtAl2024, HossjerDiazRao2022, HossjerDiazRao2026, LiuEtAl2023, RaoLiuDiaz2024, ThorvaldsenHossjer2020, ThorvaldsenHossjer2023, ThorvaldsenHossjer2024, ThorvaldsenOhrstromHossjer2024, ZhouEtAl2023}.

 \begin{figure}[t]
    \centering
    \includegraphics[width=0.3\textwidth]{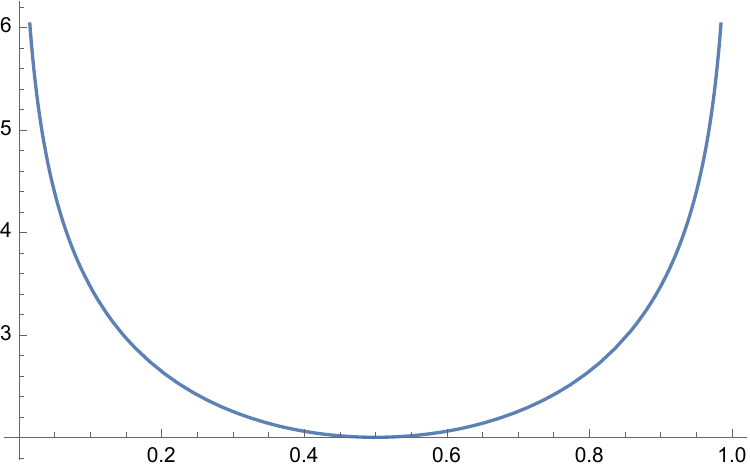}
    	\caption{{\footnotesize Total information $\calH(X)$ of $X\sim\textrm{Ber}(p)$. Logs taken in base 2.}}\label{FigTI}
\end{figure}  

\section{Conserved active information}

We propose here {\bf conserved active information}. To introduce it, we start with a simple question: If AIN can be negative, why is KL nonnegative? To answer this question, we first define the \textbf{total information} of a random variable $X$ with law $\bfP_1$ as
\begin{align}\label{TI}
	\calH(X) = \calH(\bfP_1)&\defeq\int_\calX -\log \d\bfP_1(x).
\end{align}
We further assume that $\calH$ takes vales on $[0,\infty]$, implying that $\log 0$ is well defined. When $X$ is discrete, the RHS of \eqref{TI} becomes $-\sum_{x \in \calX} \log p_1(x)$; when $X$ is absolutely continuous, it becomes $-\int_\calX \log p_1(x) \d x$. As its name indicates, $\calH$ quantifies the total amount of information in a given system.

\begin{figure}[t]
    \centering
    \includegraphics[width=0.3\textwidth]{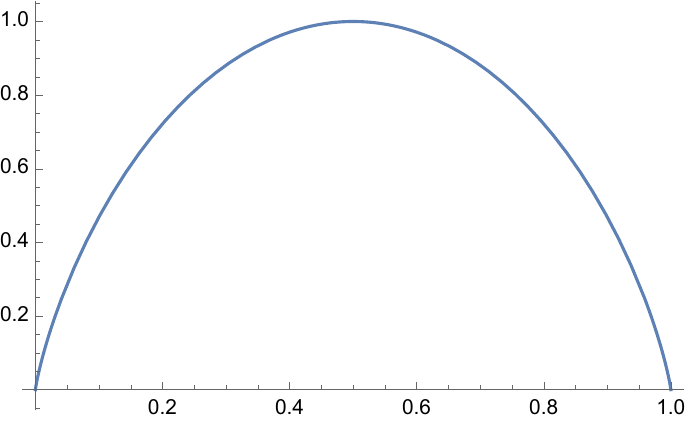}
    	\caption{{\footnotesize Entropy $H(X)$ of $X\sim\textrm{Ber}(p)$. Logs taken in base 2}}\label{FigEntropy}
\end{figure}

\begin{example}\lb{BerEx}
	Consider a Bernoulli r.v.~$X$ in $\calX=\{0,1\}$ with probability of success $p = \bfP_1(1)$ and entropy $H(X)$. Then
	\begin{align}\lb{HH}
    \begin{aligned}
		\calH(X) &= -\log p - \log(1-p),\\
		H(X) &= -p\log p -(1-p)\log(1-p)
    \end{aligned}
	\end{align}
	It is easy to verify that when $p=0.5$, $\calH(X)$ is minimized (see Fig.~\ref{FigTI}), whereas the entropy $H(X)$ is maximized (see Fig. \ref{FigEntropy}). In base 2, $\calH(1/2) = 2$ and $H(1/2) = 1$. On the other hand, as $p$ approaches 0 or 1, $\calH(X)$ increases to infinity, while $H(X)$ decreases to 0. In more detail, for $p \ll 1$,
	\begin{align}\lb{HH2}
    \begin{aligned}
		\calH(X) &\approx -\log p \gg 0,\\
		H(X) &\approx -p\log p \ll 1,
    \end{aligned}
	\end{align}
	revealing that the contribution of the second terms in \eqref{HH} becomes negligible. Still, it is also clear from \eqref{HH2} that the small weight of the first term of $H(X)$ tames the strength of the first term in $\calH(X)$.
 \end{example}

Example \ref{BerEx} shows that total information and entropy move in opposite directions. When uncertainty is maximal, entropy is maximized, while total information is minimized; when certainty is maximal, entropy is minimized, while total information is maximized. 

The previous setting can be naturally extended to {\bf conserved active information}, defined as
\begin{align}\lb{RTI}
\begin{aligned}
	\cain &= \cain(\bfP_2, \bfP_1) = \cain(X_2, X_1) \\
    &\defeq \calH(X_1) - \calH(X_2) = \int_\calX \log \frac{p_2(x)}{p_1(x)} \d\mu,
\end{aligned}
\end{align}
where $X_i \sim \bfP_i$ ($i\in\{1,2\}$), $\bfP_2 \ll \bfP_1$, and $\mu, p_2, p_1$ are as defined in \eqref{AIN2}. We assume $\cain$ takes values in $[-\infty, \calH(X_1)]$. Moreover, the absolute continuity of $\bfP_2$ with respect to $\bfP_1$, together with the assumption of $\log(0/0) = 0$ in \eqref{AIN}, ensures that $\cain$ is well-defined.

When $X_1, X_2$ are discrete, the RHS of \eqref{RTI} becomes $-\sum_{x \in\calX}\log [p_2(x)/p_1(x)]$; when $X_1, X_2$ are absolutely continuous, the RHS of \eqref{RTI} becomes $-\int_\calX \log [p_2(x)/p_1(x)] \d x$. Then $\cain$ is the additional information needed in the system to replace $X_1\sim \bfP_1$ with $X_2\sim \bfP_2$. Therefore, when $\calH(X_2) \le \calH(X_1)$, then $\cain \ge 0$, implying that the system's information is sufficient to shift from $\bfP_1$ to $\bfP_2$. On the other hand, when $\calH(X_2) > \calH(X_1)$, then $\cain < 0$, and conservation of information implies that exogenous information is required to obtain the new system induced by $X_2$. The definition of $\cain$ and Pinsker inequality \eqref{TV} directly imply the following Lemma:

\begin{lemma} Let $\calX =\{1, \ldots, N\}$ be a finite space and define $X_1\sim \calU(N)$ as a uniform random variable in $\calX$. Then, for any $X_2\sim\bfP_2$ fully  supported in $\calX$, we have
    \begin{enumerate}
    \item  $N\cdot\KL{\bfP_1}{\bfP_2} = \cain(\bfP_2,\bfP_1)$.
    \item $\mathrm{TV}(\bfP_2,\bfP_1) \le \sqrt{\cain(\bfP_2,\bfP_1)/(2N)}$.
    \end{enumerate}
\end{lemma}

As before, a brief example will shed considerable light on the interpretation of $\cain$.

\begin{figure}[t]
    \centering
    \includegraphics[width=0.3\textwidth]{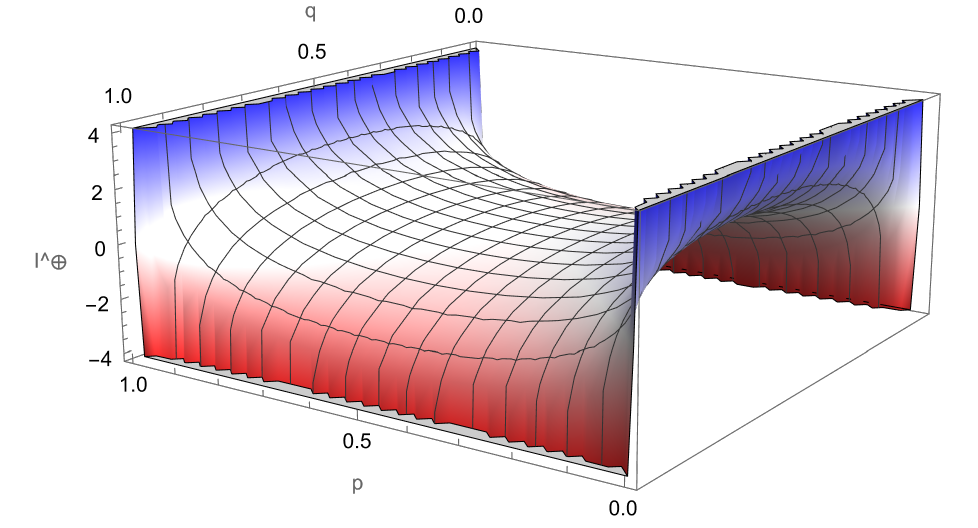}
    	\caption{{\footnotesize $\cain(\bfP_2,\bfP_1)$ when $\bfP_1 \sim\mathrm{Ber}(p)$ and $\bfP_2 \sim\mathrm{Ber}(q)$. The region in faded blue represents positive CAI. The region in faded red represents negative CAI.}}\label{FigRTI}
\end{figure}

\begin{example}\lb{CainBer}
Consider the space $\calX = \{0,1\}$, and two Bernoulli random variables $X_1, X_2$ with success probabilities $\bfP_1(\{X_1=1\}) = p$ and $\bfP_2(\{X_2=1\}) = q$, respectively (with $p,q\in(0,1)$). Figure \ref{FigRTI} shows that 
\begin{align}\lb{CBer}
    \cain = \log \frac{q}{p} + \log \frac{1-q}{1-p}
\end{align}
is a saddle. Three simple scenarios can be discerned:
\begin{enumerate}[label=\textbf{C\arabic*}]
    \item\lb{Ceq} \ \ If $p \in \{q, 1-q\}$, then $\cain=0$.
    \item\lb{Cup} \ \ If $p \ne 0.5 = q$, then $\cain > 0$. 
    \item\lb{Cdown} \ \ If $p = 0.5 \neq q$, then $\cain < 0$. 
\end{enumerate}
In scenario \ref{Ceq}, the system's total information remains constant. In scenario \ref{Cup}, a somewhat ordered system decays to maxent, implying that the system redistributes its available information without external input. In scenario \ref{Cdown}, the system is initially in maximum entropy (maxent), and any update to $\bfP_2$ orders the entire system, requiring external input. In other words,  CAI increases or decreases as the system becomes more or less ordered. 
\end{example}

\begin{figure}[t]
    \centering
    \includegraphics[width=0.37\textwidth]{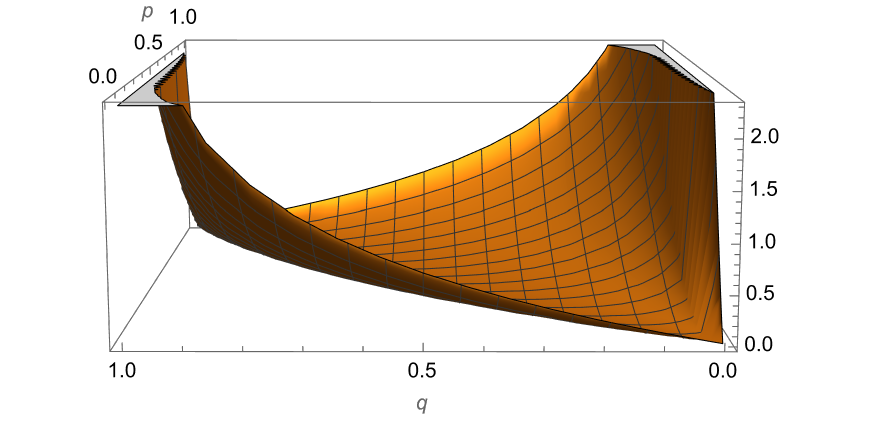}
    	\caption{{\footnotesize $\KL{\bfP_1}{\bfP_2}$} when $\bfP_1 \sim\mathrm{Ber}(p)$ and $\bfP_2 \sim\mathrm{Ber}(q)$.}\label{FigKL}
\end{figure} 

How can KL be nonnegative if $\ain$ and $\cain$ can be negative? (Figure \ref{FigKL}). Again, let's consider the simplest case of two Bernoulli random variables: $X_2 \sim \mathrm{Ber}(0.5 +\epsilon)$ and $X_1 \sim \mathrm{Ber}(0.5)$, with $\epsilon \in (0, 0.5]$. Then
\begin{align}\lb{CvsKL}
\begin{aligned}
	\cain(\bfP_2, \bfP_1) &= \log \frac{0.5-\epsilon}{0.5} + \log \frac{0.5+\epsilon}{0.5}\\ 
	 \KL{\bfP_2}{\bfP_1}  &= (0.5-\epsilon)\log \frac{0.5-\epsilon}{0.5} + (0.5+\epsilon)\log  \frac{0.5+\epsilon}{0.5}.
\end{aligned}
\end{align}

Example \ref{CainBer} and Fig. \ref{FigRTI} show that $\cain <0$ in \eqref{CvsKL}. Moreover, the first term of $\cain$ is negative, whereas the second is positive. Therefore, the first term determines the sign of $\cain$. However, KL is positive not because all its terms are positive (the weighted first term is not), but because the positive, weighted second term outweighs the weighted first term. In fact, as $\epsilon \uparrow 0.5$, $\cain$ decreases to $-\infty$ because its first term decreases to $-\infty$, whereas KL approaches $\log 2$ because its weighted first term approaches 0.  
 
As Example \ref{CainBer} makes clear, $\cain$ reveals information hidden from KL, with strong implications for the conservation of information.  This is made explicit in the theorem below.

\begin{theorem}[Regimes of CAI]\lb{Regimes} 
In the setting of Example \ref{CainBer}, the following regimes hold:
\begin{enumerate}[label=\textbf{T\arabic*}]
	\item\lb{T1} for $p=1/2$,
	
	if $q \ne 1/2$, then $\cain < 0$.
	\item\lb{T2} for $p < 1/2$,
	
	if $q>1-p$ or $q<p$, then $\cain < 0$. If $p < q < 1-p$, then $\cain >0$.
	
	\item\lb{T3} for $p > 1/2$,
	
	if $q>p$ or $q<1-p$, then $\cain < 0$. If $1-p < q < p$, then $\cain >0$.
	
	\item\lb{T4} for any $p \in(0,1]$,
	
	if $q \in \{p,1-p\}$, then $\cain=0$.
\end{enumerate}
\end{theorem}

\begin{proof}
    \begin{align*}
        \cain(\bfP_2,\bfP_1) = \log\frac{q(1-q)}{p(1-p)},
    \end{align*}
    which proves directly \ref{T1} and \ref{T4}. As for \ref{T2}, after some algebraic manipulation, we obtain that,
    \begin{align}\lb{EqTh}
        \log\frac{q(1-q)}{p(1-p)} < 0 &\Leftrightarrow (q-p)(1-q-p) < 0.
    \end{align}
    Thus, the two terms on the RHS of \eqref{EqTh} must have opposite signs. Therefore, if $q-p>0$, \eqref{EqTh} holds iff $1-p <q$. And, if $q-p < 0$, \eqref{EqTh} holds iff $q<p<1/2< 1-p$. The proof of \ref{T3} is analogous.
\end{proof}


Theorem \ref{Regimes} has the following important corollary:

\begin{corollary}\lb{Co:regimes}
    Under the conditions of Theorem \ref{Regimes}, 
    \begin{enumerate}
    	\item for $p< 1/2$,
	
	if $\ain< 0$, then $\cain < 0$. On the other hand, if $\ain >0$, then $\cain < 0$ only when $q>1-p$. Otherwise, if $\ain>0$ and $p<q<1-p$, then $\cain > 0$.
	
	\item for $p>1/2$,
	
	if $\ain > 0$, then $\cain < 0$. On the other hand, if $\ain <0$, then $\cain <0$ only when $q< 1-p$. Otherwise, if $\ain>0$ and $1- p<q<p$, then $\cain > 0$.
	
	\item for $p=1/2$, $\cain <0$, irrespective of $\ain$.

    \end{enumerate}
    
\end{corollary}

Corollary \ref{Co:regimes} reveals properties that hold beyond the Bernoulli case and that cannot be discerned by looking only at KL. Consider two probability measures $\bfP_1$ and $\bfP_2$ on a measurable space $(\calX, \calF)$ and a specification $\sfT \in \calF$. If $\cain  \ge 0$, the probability space $(\calX, \calF, \bfP_2)$ can be obtained by redistributing the information available in $(\calX, \calF, \bfP_1)$; moreover, if at the same time $\ain(\sfT) >0$, the information required to produce $\sfT$ can be obtained by such a redistribution. For instance, when $\calX = \{0,1\}$ and $\bfP_1(\sfT) < 1/2$, Theorem \ref{Regimes}.\ref{T4} shows that it is possible to increase $\bfP_2(\sfT)$ up to $1-\bfP_1(\sfT) > 1/2$, making $\ain(\sfT) > 0$ using only the total information available in $(\calX, \calF, \bfP_1)$; this is particularly surprising if $0 < \bfP_1(\sfT) \ll 1/2$, since $\bfP_2(\sfT)$ can become very large by redistributing the information in $(\calX, \calF, \bfP_1)$ alone. We call this scenario the reducible regime.

On the other hand, if $\ain(\sfT) > 0$ and $\cain < 0$, the probability space $(\calX, \calF, \bfP_2)$ contains more information than is available in $(\calX, \calF, \bfP_1)$. Therefore, there is no way to increase the likelihood of $\sfT$ from $\bfP_1(\sfT)$ to $\bfP_2(\sfT)$ by redistributing the information in $(\calX, \calF, \bfP_1)$. This is particularly true when $\sfT$ is a strict subset of $\calX$ that is observed and $\bfP_1(\sfT) < 1$, in which case $\bfP_2(\sfT) = 1$ and $\cain = \int_\sfT \ain \d\mu + \int_{\sfT^c}\ain \d\mu = -\infty$, since in this scenario we know that $(\calX,\calF, \bfP_1)$ did not possess the information required to produce $\sfT$. We call this scenario the emergent regime. Accordingly, for any probability space, we define a specification as emergent if it is in this regime. The possible regimes are summarized in Table \ref{tab:regimes}.

\begin{definition}[Emergence]\lb{Emergence}
	We say that a specification $\sfT \subset \calX$ is emergent if and only if $\ain(\sfT)>0$ and $\cain < 0$.
\end{definition}	

\begin{remark}
	Theorem \ref{Regimes} shows that, despite its name, $I_2$ in \eqref{AIN} and \eqref{AIN2} is only {\it exogenous} in the emergent regime. However, the name is not warranted in the reducible regime.
\end{remark}

\begin{table*}[t]
\centering
\caption{\lb{tab:regimes} Conserved active information regimes}
\begin{tabular}{c|c|c|l}
\hline
\textbf{Regime} & \textbf{Condition}                 & \textbf{Bernoulli example}     & \multicolumn{1}{c}{\textbf{Description}}     \\ \hline
\multirow{2}{*}{Reducible} & $\ain(\sfT) < 0$ and $\cain \ge 0$ & $p > 1/2$, $q\in [1-p,p]$ & Specification harder, system more disordered \\ \cline{2-4} 
                & $\ain(\sfT) > 0$ and $\cain \ge 0$ & $p < 1/2$, $q \in [p, 1-p]$ & Specification easier, system more disordered \\ \hline
\multirow{2}{*}{Emergent}  & $\ain(\sfT) < 0$ and $\cain < 0$   & $p > 1/2$, $q< [0,1-p)$   & Specification harder, system more ordered    \\ \cline{2-4} 
                & $\ain(\sfT) > 0$ and $\cain < 0$   & $p< 1/2$, $q > (1-p,1]$     & Specification easier, system more ordered    \\ \hline
\end{tabular}
\end{table*}

\begin{example}[Markov chains \cite{LevinPeres2017}]\lb{MC}
    Let $\calG = (\calX, E)$ be a connected graph with finite vertex set $\calX$, and assume it is $d$-regular, meaning every vertex has degree $d$. The associated random walk has transition matrix
    \begin{equation*}
        P(x,y) =
        \begin{cases}
            1/d & \mathrm{if \ } (x,y) \in E \\
            0   & \mathrm{otherwise}.
        \end{cases}
    \end{equation*}
    Connectivity and regularity of the graph imply that the random walk is irreducible and has a unique stationary distribution $\bfP_2$ which is uniform over $\calX$. Let $\sfT\subset \calX$ be a specification such that $\abs{\sfT} < 1/2$, and consider an initial distribution $\bfP_1$ such that $0 < \bfP_1(\sfT) \ll \abs{\sfT}/\abs{\calX}$. Then, the stationary distribution of the random walk satisfies $q= \bfP_2(\sfT) = \abs{\sfT}/\abs{\calX}$, greatly increasing the probability of the target through the chain's neutral dynamics. This illustrates the reducible regime $p<q<1-p$ in Theorem \ref{Regimes}, showing that the probability of the specification can be increased through natural information redistribution within the system.
\end{example}

\begin{example}[Cosmological fine-tuning \cite{LewisBarnes2016}]
    Cosmological fine-tuning (FT) holds that some constants of nature must lie within an interval $\ell \subset \calX \subset \bbR$ for carbon-based life to exist \cite{Carter1974}. Let $\sfT \subset \calX$ be the specification ``We observe a universe that exists and permits life.'' Let $x$ denote the value of a particular constant of nature, and assume that $x$ is a parameter of a model $\bfQ\giventhat{\sfT}{x}$ that gives the probability of observing a life-permitting universe. According to \eqref{T}, this corresponds to $\sfT$ when the specification function $f(x)=\1_\sfT(x)$ is binary and satisfies $f(x)=1$ if the value $x$ of the constant permits a universe with life ($f(x)=0$ otherwise). 
    
The tuning probability of $\sfT$ is obtained by treating $x$ as an observation of a continuous random variable $X\sim \bfP$, where $\bfP$ is in maxent, and averaging $\bfQ\giventhat{\sfT}{x}$ with respect to $x$, i.e., $\bfQ(\sfT;\xi) = \int_\calX \bfQ\giventhat{\sfT}{x}\d\bfP(x;\xi)$, where $\bfQ\giventhat{\sfT}{x}$ is the likelihood. If $\sfT = \ell$, then $\bfQ\giventhat{\sfT}{x} = \1_\sfT(x)$, and we obtain $\bfQ(\sfT;\xi) = \bfP(\sfT ;\xi)$. To determine the degree of tuning, we maximize $\bfP(\sfT;\xi)$ with respect to the hyperparameter $\xi$ as $\xi$ varies over a finite-dimensional space $\Xi$. That is, the final degree of tuning equals $\bfP_1(\sfT) \defeq \sup_{\xi\in\Xi} \bfP(\sfT;\xi)$. Since $\sfT$ is observed, this induces a second distribution $\bfP_2$ s.t.~$\bfP_2(\sfT)=1$. Then $\bfP_0(\sfT) < \delta \ll 1$ is equivalent to $\ain(\sfT) > -\log \delta \gg 0$, in which case we infer that $X$ is fine-tuned to level $\delta$ for the family of distributions $\frakF \defeq \{\bfP(\sfT;\xi)\}_{\xi \in \Xi}$. 

In this scenario, it has been shown that the gravitational constant and the critical density of the universe are fine-tuned. In contrast, other constants, such as the Higgs vacuum expectation value and the amplitude of the primordial fluctuations, are not \cite{DiazHossjer2022, DiazHossjerMarks2021, DiazHossjerMarks2023, DiazHossjerMathew2024}. Therefore, by Corollary \ref{Co:regimes}, the presence of fine-tuning, $\bfP_1(\sfT) \ll 1$ and $\bfP_2(\sfT) = 1$, implies that the universe itself is emergent.
\end{example}

\section{Conclusion}

In this paper, we have introduced total information and CAI. We have highlighted their properties and contrasted them with key properties of AIN and KL, which are rarely discussed in the literature. Since neither AIN nor KL can describe changes in the system's total information, $\cain$ provides a necessary criterion for determining whether a specification $\sfT$ requires the incorporation of system-specific knowledge.

The implications of CAI are fascinating. Consider the binary case in Example \ref{CainBer}, Theorem \ref{Regimes}, and Corollary \ref{Co:regimes}, with specification $\sfT = \{1\} \subset \calX$. On the one hand, the two cases of the reducible regime should not be surprising, since focusing on $\sfT$ (with baseline probability $p \le 1/2$) to make it more probable is equivalent to focusing on $\sfT^c$ (with baseline probability $1-p$) to make it less probable, and this should not be controversial. On the other hand, the emergent regime shows that $q> 1-p$ cannot be achieved by redistributing the information available in the system. Similar considerations to those in the reducible regime reveal that the order achieved by making $\sfT$ highly likely ($0<p<1/2<1-p<q$) is equivalent to the order achieved by making $\sfT^c$ highly unlikely ($1-q< p$), linking regimes 3 and 4 in Table \ref{tab:regimes}. In this scenario, external information is required to produce the specification $\sfT$, making the system $(\calX,\calF, \bfP_1)$ open. 

This work can be extended in several ways. First, a natural extension of CAI is to explore its thermodynamic implications, where analogies between information theory and thermodynamics are particularly fruitful \cite{Jaynes1957a, Jaynes1957b}. Informally, by considering {\it energy} in our framework rather than {\it information}, we can draw direct parallels that illuminate far-from-equilibrium processes central to complex systems. In this scenario, total information is akin to total energy in a closed system. Negative CAI signals the infusion of external work (analogous to problem-specific knowledge in a search problem), reducing global disorder, much like an engine extracting work to decrease local disorder while increasing overall entropy \cite{Shannon1948, Landauer1961}. Positive CAI, conversely, reflects internal redistribution without net gain, mirroring dissipative processes in which entropy rises spontaneously, as in Example \ref{MC}.

Second, a compelling extension of CAI concerns its potential role in evaluating artificial intelligence (AI) systems, particularly large language models (LLMs), and in advancing understanding of artificial general intelligence (AGI). In search and optimization, LLMs leverage vast pre-trained knowledge to enhance performance on specific tasks, effectively increasing active information by biasing toward relevant targets \cite{DiazEtAl2025, HossjerDiazRao2022, HossjerDiazRao2026}. We conjecture that this often occurs within the reducible regime identified in the second row of Table \ref{tab:regimes}, where $\ain(\sfT)>0$ and $\cain > 0$, indicating that the system becomes more disordered overall. Here, gains in active information stem from internal redistribution of probabilities---drawing on the programmer's embedded knowledge (e.g., training data curation and architecture design)---rather than from a net injection of order. In contrast, true AGI could be characterized by the machine's autonomous ability---not the programmer's---to transition into the emergent regime ($q>1-p$ and $\cain <0$), where it imposes genuine order on the entire space by injecting problem-specific ``insights'' without external scaffolding. This jump to negative $\cain$ would signify self-generated information, akin to emergent creativity or adaptation in unexplored landscapes.

Third, in statistical estimation, consider a population $\calX$ and a subset $\sfT \subset \calX$ of infected individuals with a disease. The true prevalence is $p=\bfP_1(\sfT)=\abs{\sfT}/\abs{\calX}$. An estimator introduces bias, yielding $q = \bfP_2(\sfT)$ \cite{DiazEtAl2025, HossjerEtAl2024}. Here, active information $\ain(\sfT) = \log(q/p)$ quantifies the magnitude and direction of bias: positive for overestimation, negative for underestimation, and 0 for unbiasedness. We believe the regimes delineated in Table \ref{tab:regimes} provide an objective framework for assessing permissible bias. Specifically, the reducible regime ($p < q < 1/2 < 1-p$) bounds how much, under unbiased sampling, an estimator can inflate $q$ without violating conservation principles by injecting excessive external knowledge. For rare events where $p \ll 1/2$, the interval $(p, 1/2)$ is vast, allowing substantial overestimation ($q \gg p$) while keeping $\cain > 0$. Consistent with conservation of information, estimators may appear effective locally but degrade globally, leading to overconfidence in rare-event detection. In contrast, crossing into the emergent regime signals that the estimator has imposed genuine order, potentially through unaccounted-for external inputs (e.g., biased sampling or overfitting), thereby violating conservation of information unless justified by knowledge of problem-specific structure. This framework thus offers a diagnostic tool for bias in prevalence studies: compute $\cain$ to determine whether the observed bias is mild (permissible redistribution) or strong (indicating the need for scrutiny of hidden knowledge).

Fourth, this implies that the reducible regime gives traditional evolutionary accounts {\it a lot of} leverage for an evolutionary process and a given target $\sfT$. Additionally, the emergent regime shows that the traditional accounts must be supplemented with information from higher biological scales (atoms and ions, molecules, networks, organelles, cells, tissues, organs, organ systems, and organisms). This is in line with Denis Noble's biological relativity, according to which each scale is open and receives information from higher scales \cite{Noble2017}.  For instance, a specification $\sfT$ at the organelle scale in the emergent regime would imply that it received information from a higher scale, from cells to organisms.
\bibliography{/Users/daangapa/Documents/Research/daangapaBibliography.bib}

\end{document}